%% file: sample-sigconf.tex
\documentclass[sigconf]{acmart}

\usepackage{silence}
\usepackage{hyperref}

\renewcommand\footnotetextcopyrightpermission[1]{} 
\usepackage{subcaption}
\usepackage[english]{babel}
\usepackage[utf8]{inputenc}
\usepackage{algorithm}
\usepackage[noend]{algpseudocode}
\usepackage{multirow}
\usepackage{mathtools}
\usepackage{booktabs} 

\usepackage[small,compact]{titlesec} 
\titlespacing\section{0pt}{4pt}{0pt}
\titlespacing\subsection{0pt}{4pt}{0pt}

\setcopyright{none}

\begin{document}
\title{Understanding the Logical and Semantic\\ Structure of Large Documents}

\author{Muhammad Mahbubur Rahman}
\affiliation{%
  \institution{University of Maryland, Baltimore County}
  \city{Baltimore} 
  \state{Maryland} 
  \postcode{21250}
}
\email{mrahman1@umbc.edu}

\author{Tim Finin}
\affiliation{%
  \institution{University of Maryland, Baltimore County}
  \city{Baltimore} 
  \state{Maryland} 
  \postcode{21250}
}
\email{finin@umbc.edu}


\begin{abstract} \noindent
Current language understanding approaches focus on small documents, such as newswire articles, blog posts, product reviews and discussion forum entries. Understanding and extracting information from large documents like legal briefs, proposals, technical manuals and research articles is still a challenging task. We describe a framework that can analyze a large document and help people to know where a particular information is in that document. We aim to automatically identify and classify semantic sections of documents and assign consistent and human-understandable labels to similar sections across documents. A key contribution of our research is modeling the logical and semantic structure of an electronic document. We apply machine learning techniques, including deep learning, in our prototype system. We also make available a dataset of information about a collection of scholarly articles from the \textit{arXiv} eprints collection that includes a wide range of metadata for each article, including a table of contents, section labels, section summarizations and more. We hope that this dataset will be a useful resource for the machine learning and NLP communities in information retrieval, content-based question answering and language modeling.

\end{abstract}

%
%



\maketitle

\input{samplebody-conf}

\bibliographystyle{ACM-Reference-Format}
\bibliography{sample-sigconf} 

\end{document}

%% file: samplebody-conf.tex
\section{Introduction} 

Understanding and extracting of information from large documents such as reports, business opportunities, academic articles, medical documents and technical manuals poses challenges not present in short documents. And state of the art natural language processing approaches mostly focus on short documents such as newswire articles, dialogs, blog posts, product reviews and discussion forum entries. One of the key challenges in the processing of large documents is sectioning different parts of a document. The reason behind this challenge is that large documents are complex, may be unstructured and noisy with different formats.

Document understanding depends on a reader's own interpretation, where a document may structured, semi-structured or unstructured. Usually a human readable document has a physical layout and logical structure. A document contains sections. Sections may contain a title, section body or a nested structure. Sections are visually separated components by a section break such as extra space, empty line or a section heading for the latter section. A section break signals to a reader the changes of concept, mood, tone and emotion. The lack of proper transition from one section to another section may raise the difficulty to understand the document. 

Understanding large multi-themed documents presents additional challenges as these documents are composed of a variety of sections discussing diverse topics. Some documents may have a table of contents whereas others may not. Even if a table of contents is present, mapping it across the document is not a straightforward process. Section and subsection headers may or may not be present in the table of contents. If they are present, they are often inconsistent across documents even within the same vertical domain.

Most of the large documents such as business documents, health care documents and technical reports are available in PDF format. This is because of the popularity and portability of PDF file over different types of machines. But PDF is usually rendered by various kind of tools such as Microsoft Office, Adobe Acrobat and Open Office. All of these tools have their own rendering techniques. Moreover, content is written and formatted by people. All of these factors make PDF documents very complex with text, images, graphs and tables. 

Semantic organization of sections, subsections and sub-subsections of PDF documents across all vertical domains are not the same. For example, a business document has a completely different structure from a user manual. Even research articles from computer science and social science have completely different structures. Social science articles have methodology sections where as computer science articles have approach sections. Semantically these two sections should be the same.

\begin{figure}
\includegraphics[height=2.0in, width=3.2in]{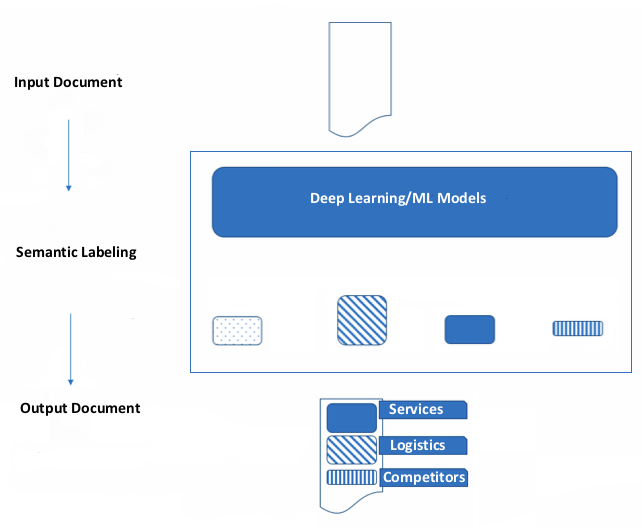}
\caption{A High Level System Work-flow\label{fig:highlevelSystemwork-flow}}
\end{figure}

\begin{figure}
\includegraphics[height=1.0in, width=2.3in]{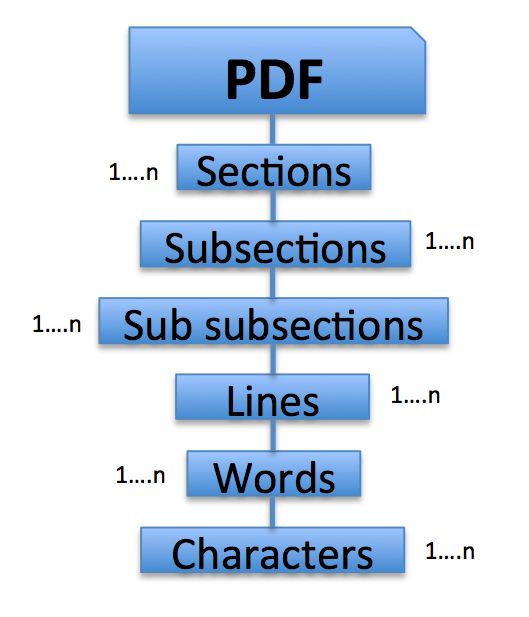}
\caption{Logical Model of a PDF Document\label{fig:document_model}}
\end{figure}

We intend to section large and complex PDF documents automatically and annotate each section with a semantic and human-understandable label. Figure \ref{fig:highlevelSystemwork-flow} shows the high level system work-flow of our framework. The framework takes a document as input, extracts text, identifies logical sections and labels them with semantically meaningful names. The framework uses layout information and text content extracted from PDF documents. A  logical model of a PDF document is given in Figure \ref{fig:document_model}, where each document is a collection of n sections and a section is a collection of subsections and so on. 

Identifying a document's logical sections and organizing them into a standard structure to understand the semantic structure of a document will not only help many information extraction applications but also enable users to quickly navigate to sections of interest. Such an understanding of a document's structure will significantly benefit and inform a variety of applications such as information extraction and retrieval, document categorization and clustering, document summarization, fact and relation extraction, text analysis and question answering. People are often interested in reading specific sections of a large document and hence will find semantically labeled sections very useful. It will help people simplify their reading operations as much as possible and save valuable time. 

One might be confused that document sectioning and semantic labeling are the same as document segmentation \cite{antonacopoulos2013icdar}, but these are distinct tasks. Document segmentation is based on a scanned image of a text document. Usually a document is parsed based on raw pixels generated from a binary image. We use electronic documents such as PDFs generated from Word, LaTeX or Google Doc and consider different physical layout attributes such as indentation, line spaces and font information. 

One might also confuse semantic labeling with rhetorical or coherence relations of text spans in a document. Rhetorical Structure Theory (RST) \cite{mann1988rhetorical,taboada2006rhetorical} uses rhetorical relations to analyze text in order to describe rather than understand them. It finds coherence in texts and parses their structure. This coherence is helpful for identifying different components of a text block, but we aim to understand the text blocks in order to associate a semantic meaning.


\section{Background}
This section provides necessary background on our research and includes definitions required to understand the work.

\subsection{Sections}
A section can be defined in different ways. In our paper, we define a section as follows. \\
$S$ $=$ a set of \textit{paragraphs}, \;$P$ ; where number of paragraphs is $1$ \textit{to} $n$  \\
$P$ $=$ a set of \textit{lines},\;$L$ \\
$L$ $=$ a set of \textit{words},\;$W$  \\
$W$ $=$ a set of \textit{characters}, \;$C$ \\
$C$ $=$ all character set\\
$D$ $=$  \textit{digits} $|$ \textit{roman numbers} $|$ \textit{single character}  \\
$LI$ $=$ a set of \textit{list items} \\
$TI$ $=$ an entry from a table \\
$Cap$ $=$ \textit{table caption} $|$ \textit{image caption} \\
$B$ $=$ characters are in \textit{Bold} \\
$LFS$ $=$ characters are in \textit{larger font size} \\
$HLS$ $=$ higher line \textit{space} \\\\
$\textit{Section Header}$ $=$  \begin{math} $l$ \; \subset \; $L$  \end{math}  where $l$ often starts with  \begin{math} $d$\; \in \;$D$ \end{math}  \textbf{And} \begin{math} $l$ \;\notin\; \{$TI$,\; $Cap$\} \end{math} \textbf{And} \begin{math} usually\;$l$ \;\in\; $LI$ \end{math}  \textbf{And} generally \begin{math} $l$ \;\subset\; \{$B$,\; $LFS$,\;$HLS$\} \end{math} \\\\
$\textit{Section}$ $=$ \begin{math} $s$ \; \subset \; $S$ \end{math} followed by a $\textit{Section Header}$. 
\subsection{Documents}
Our work is focused on understanding the textual content of PDF documents that may have a few pages to a few hundred pages.  We consider those with more than ten pages to be "large" document. It is common for them to have page headers, footers, tables, images, graphics, forms and mathematical equation. Some examples of large documents are business documents, legal documents, technical reports and academic articles. 

\subsection{Document Segmentation}
Document segmentation is a process of splitting a scanned image from a text document into text and non-text sections. A non-text section may be an image or other drawing. And a text section is a collection of machine-readable alphabets, which can be processed by an OCR system. Usually two main approaches are used in document segmentation, which are geometric segmentation and logical segmentation. According to geometric segmentation, a document is split into text and non-text based on its geometric structure. And a logical segmentation is based on its logical labels such as header, footer, logo, table and title. The text segmentation is a process of splitting digital text into words, sentences, paragraphs, topics or meaningful sections. In our research, we are splitting digital text into semantically meaningful sections with the help of geometrical attributes and text content.

\subsection{Document Structure}
A document's structure can be defined in different ways. In our research, documents have a hierarchical structure which is considered as the document's logical structure. According to our definition, a document has top-level sections, subsections and sub-subsections. Sections start with a section header, which is defined in the earlier part of the background section. A document also has a {\em semantic structure}. An academic article, for example, has an abstract followed by an introduction whereas a business document, such as an RFP, has deliverables, services and place of performance sections. In both the logical and semantic structure, each section may have more than one paragraph. 

\section{Related Work}
Identifying the structure of a scanned text document is a well-known research problem. Some solutions are based on the analysis of the font size and text indentation \cite{bloomberg1996document, mao2003document}. Song Mao et al. provide a detailed survey on physical layout and logical structure analysis of document images \cite{mao2003document}. According to them, document style parameters such as size of and gap between characters, words and lines are used to represent document physical layout. 
Algorithms used in physical layout analysis can be categorized into three types: top-down, bottom-up and hybrid approaches. Top-down algorithms start from the whole document image and iteratively split it into smaller ranges. Bottom-up algorithms start from document image pixels and cluster the pixels into connected components such as characters which are then clustered into words, lines or zones. A mix of these two approaches is the hybrid approach. 

The O'Gorman's Docstrum algorithm \cite{o1993document}, the Voronoi-diagram-based algorithm of Kise \cite{kise1998segmentation} and Fletcher's text string separation algorithm  \cite{fletcher1988robust} are bottom-up algorithms. Lawrence Gorman describes Docstrum algorithm using the K-nearest neighbors algorithm \cite{fukunaga1975branch} for each connected component of a page and uses distance thresholds to form text lines and blocks. Kise et al. propose Voronoi-diagram-based method for document images with a non-Manhattan layout and a skew.  Fletcher et al. design their algorithm for separating text components in graphics regions using Hough transform \cite{kiryati1991probabilistic}.
The X-Y-cut algorithm presented by Nagy et al. \cite{nagy1992prototype} is an example of the top-down approach based on recursively cutting the document page into smaller rectangular areas. A hybrid approach presented by Pavlidis et al. \cite{pavlidis1992page} identifies column gaps and groups them into column separators after horizontal smearing of black pixels.

Jean-Luc Bloechle et al. describe a geometrical method for finding blocks of text from a PDF document and restructuring the document into a structured XCDF format \cite{bloechle2006xcdf}. Their approach focuses on PDF formatted TV Schedules and multimedia meeting note, which usually are organized and well formatted. Hui Chao et al. describe an approach that automatically segments a PDF document page into different logical structure regions such as text blocks, images blocks, vector graphics blocks and compound blocks \cite{chao2004layout}, but does not consider continuous pages.  
Hervé Déjean et al. present a system that relies solely on PDF-extracted content using table of contents (TOC) \cite{dejean2006system}. But many documents may not have a TOC. Cartic Ramakrishnan et al. develop a layout-aware PDF text extraction system to classify a block of text from the PDF version of biomedical research articles into rhetorical categories using a rule-based method \cite{ramakrishnan2012layout}. Their system does not identify any logical or semantic structure for the processed document.

Alexandru Constantin et al. design PDFX, a rule-based system to reconstruct the logical structure of scholarly articles in PDF form and describe each of the sections in terms of some semantic meaning such as title, author, body text and references \cite{constantin2013pdfx}. They get 77.45 F1 score for top-level heading identification and 74.03 F1 score for extracting individual bibliographic items.
Suppawong Tuarob et al. describe an algorithm to automatically build a semantic hierarchical structure of sections for a scholarly paper \cite{tuarob2015hybrid}. Though, they get 92.38\% F1 score in section boundary detection, they only detect top-level sections and settle upon few standard section heading names such as ABS (Abstract), INT (Introduction) and REL (Background and Related Work). But a document may have any number of section heading names.  

Most previous work focuses on image documents, which are not similar to the problem we are trying to solve. Hence, their methods are not directly applicable to our research. Some research covers scholarly articles considering only the top-level sections without any semantic meaning. Our research focuses on any type of large document including academic articles, business documents and technical manuals. Our system understands the logical and semantic structure of any document and finds relationship between top-level sections, subsections and sub-subsections. 

\section{System Architecture and Approach}
In this section, we describe the system architecture of our framework. We explain our approaches and algorithms in detail. We also show the input and output of our framework.

\subsection{System Architecture}
\begin{figure*}
\includegraphics[height=3.0in, width=5.3in]{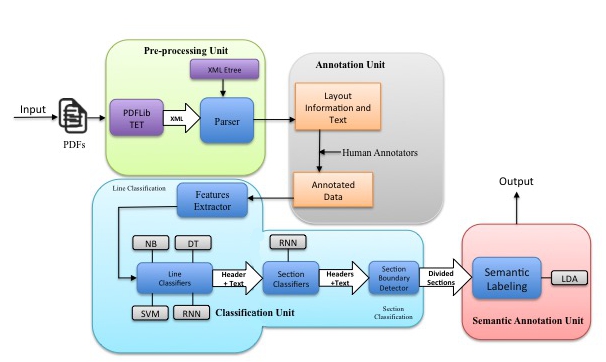}
\caption{A High Level System Architecture\label{fig:highlevelSystemarchitecture}}
\end{figure*}

Our system is organized as a sequence of units, including a Pre-processing, Annotation, Classification and Semantic Annotation units, as shown in figure \ref{fig:highlevelSystemarchitecture}.

\subsubsection{Pre-processing Unit}
The pre-processing unit takes PDF documents as input and gives processed data as output for annotation. It uses PDFLib \cite{pdflib} to extract metadata and text content from PDF documents. It has a parser, that parses XML generated by PDFLib using XML etree. The granularity of XML is word level, which means XML generated by PDFLib from PDF document has high level descriptions of each character of a word. The parser applies different heuristics to get font information of each character such as size, weight and family. It uses x-y coordinates of each character to generate a complete line and calculates indentation and line spacing of each line. It also calculates average font size, weight and line spacing for each page. All metadata including text for each line is written in a CSV file where each row has information and text of a line.  

\subsubsection{Annotation Unit}
The Annotation Unit takes layout information and text as input from the Pre-processing Unit as a CSV file. Our annotation team reads each line, finds it in the original PDF document and annotates it as a \textit{section-header} or \textit{regular-text}. While annotating, annotators do not look into the layout information given in the CSV file. For our experiments on \textit{arXiv} articles, we extract bookmarks from PDF document and use them as gold standard annotation for training and testing as described in the experiments section. 

\subsubsection{Classification Unit}
The Classification Unit takes annotated data and trains classifiers to identify physically divided sections. The Unit has sub-units for line and section classification. The Line Classification sub unit has Features Extractor and Line Classifiers module. The Features Extractor takes layout information and text as input. Based on heuristics, it extracts features from layout information and text. Features include text length, number of noun phrases, font size, higher line space, bold italic, colon and number sequence at the beginning of a line. The Line Classifiers module implements multiple classifiers using well known algorithms such as Support Vector Machines (SVM), Decision Tree (DT), Naive Bayes (NB) and Recurrent Neural Networks (RNN) as explained in the Approach section. The output of the Line Classifiers module are \textit{section-header} or \textit{regular-text}. 
The classified section header may be \textit{top-level}, \textit{subsection} or \textit{sub-subsection} header. The Section Classifiers module of the Section Classification sub unit takes section headers as input and classifies them as \textit{top-level}, \textit{subsection} or \textit{sub-subsection} header using RNN. The Section Classification sub unit also has a Section Boundary Detector which detects the boundary of a section using different level of section headers and regular text. It generates physically divided sections and finds relationship among \textit{top-level}, \textit{subsection} and \textit{sub-subsection}. It also generates a TOC from a document based on the relationship among different levels of sections, as explained further in the Approach section. 

\begin{figure}
\includegraphics[height=1.3in, width=3.0in]{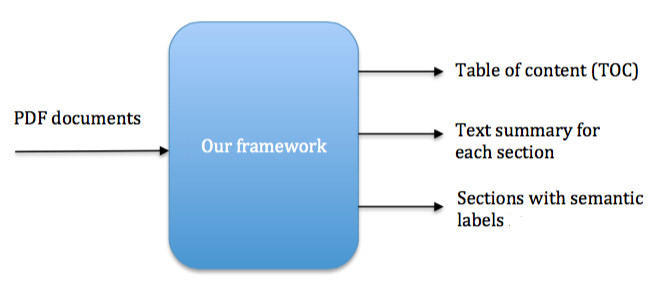}
\caption{Overall input and output of our framework\label{fig:input_output_overall}}
\end{figure}

\subsubsection{Semantic Annotation Unit}
The Semantic Annotation Unit annotates each physically divided section with a semantic name. It has a Semantic Labeling module, which implements Latent Dirichlet Allocation(LDA) topic modeling algorithm to get a semantic concept from each of the sections and annotates each section with a semantic concept understandable to people. It also applies document summarization technique using NTLK to generate a short summary for each individual section. The output are a TOC, semantic labels and a summary from each PDF document. The overall input and output of our framework are shown in figure \ref{fig:input_output_overall}.

\subsection{Approach}
In this section, we present powerful, yet simple approaches to build classifiers and models using layout information and text content from PDF documents in detail.

\subsubsection{Line Classification}
The Line Classification unit identifies each line of text as a \textit{section-header} or \textit{regular-text}. We explain our approaches for the Line Classification below. 
\subsubsection*{Features Extractor}
Given a collection of labeled text and layout information on a line, the Features Extractor applies different heuristics to extract features. We build a vocabulary from all section headers of \textit{arXiv} training data, where a word is considered if the frequency of that word is more than 100 and is not a common English word. The vocabulary size is 13371 and the top five words are "Introduction", "References", "Proof", "Appendix" and "Conclusions". The Features Extractor calculates average font size, font weight, line spacing and line indentation. It finds number of dot, sequence number, length of the text, presence of vocabulary and case of words (title case and upper case) in the text. It also generates lexical features such as the number of Noun or Noun Phrase, verb and adjective. It is common that a section header should have more Noun or Noun Phrases than other parts of speech. The ratio of verbs or auxiliary verbs should be much less in a section header. A section header usually starts with a numeric or Roman number or a single English alphabet letter. Based on all these heuristics, the Features Extractor generates 16 features from each line. These features are given in table \ref{tab:human_features}. We also use the n-gram model to generate unigram, bigram and trigram features from the text. After features generation, the Line Classifiers module uses SVM, DT, NB and RNN to identify a line as a \textit{section-header} or \textit{regular-text}.

\begin{table}
\setlength\abovecaptionskip{0pt}
  \caption{Human generated features} 
  \label{tab:human_features}
  \begin{tabular}{lp{6cm}} \\ \toprule
Feature name & 
pos\_nnp, 
without\_verb\_higher\_line\_space, 
font\_weight, 
bold\_italic, 
at\_least\_3\_lines\_upper, 
higher\_line\_space, 
number\_dot, 
text\_len\_group, 
seq\_number, 
colon, 
header\_0, 
header\_1, 
header\_2, 
title\_case, 
all\_upper, 
voc\\ \bottomrule
\end{tabular}
\end{table}

\subsubsection*{Support Vector Machines(SVM)}
Our line classification task can be considered as a text classification task where input are the layout features and n-gram from the text. Given a training data set with labels, we can train SVM models which learn a decision boundary to split the dataset into two groups by constructing a hyperplane or a set of hyperplanes in a high dimensional space. Suppose, our training dataset, 
$T$ $=$ \{$x_1$,\; $x_2$,\;....,\;$x_n$\} of text lines and their label set, $L$ $=$ \{$0$,\;$1$\} where $0$ means \textit{regular-text} and $1$ means \textit{section-header}. Each of the data points from $T$ is either a vector of 16 layout features or a vector of 16 layout features concatenated with n-gram features generated from text using $TF-IDF$ $vectorizer$. Using $SVM$, we can determine a classification model as equation \ref{eq:svm_fun} to map a new line with a class label from L. 

\begin{equation}
\label{eq:svm_fun}
  f : T \rightarrow L \;\;\;\;\;f(x) = L
\end{equation}

Here the classification rule, the function \begin{math} f(x) \end{math} can be of different types based on the chosen kernels and optimization techniques. We use LinearSVC from scikit-learn \cite{scikit-learn} which implements Support Vector Classification for the case of a linear kernel presented by Chih-Chung Chang et al. \cite{chang2011libsvm}. As our line classification task has only two class labels, we use linear kernel. We experiment with different parameter configurations for both the combine features vector and only the layout features vector. The detail of the SVM experiment is presented in the Experiments section.




\subsubsection*{Decision Tree(DT)}
Given a set of text lines, $T$ $=$ \{$x_1$,\; $x_2$,\;....,\;$x_n$\} and each line of text, $x_i$ is labeled with a class name from the label set, $L$ $=$ \{$0$,\;$1$\}, we train a decision tree model that predicts the class label for a text line, $x_i$ by learning simple decision rules inferred from either $16$ $layout features$ or $16$ $layout features$ concatenated with a number of n-gram \textit{features} generated from the text using $TF-IDF$ $vectorizer$. The model recursively partitions all text lines such that the lines with the same class labels are grouped together. 

To select the most important feature which is the most relevant to the classification process at each node, we calculate the $gini-index$. Let 
$p_1$($f$)
 and $p_2$($f$) be the fraction of class label presence of two classes  $0$: \textit{regular-text} and $1$: \textit{ section-header} for a feature $f$. Then, we have equation \ref{eq:dt_gini_1}.

\begin{equation}
\label{eq:dt_gini_1}
  \sum_{i=1}^{2}p_i(f)=1 
\end{equation}

Then, the $gini-index$ for the feature $f$ is in equation \ref{eq:dt_gini_2}.

\begin{equation}
\label{eq:dt_gini_2}
  G(f) = \sum_{i=1}^{2}p_i(f)^2
\end{equation}

For our two class line classification task, the value of $G$($f$) is always in the range of (1/2,1). If the value of $G$($f$) is high, it indicates a higher discriminative power of the feature $f$ at a certain node. 

We use decision tree implementation from scikit-learn \cite{scikit-learn} to train a decision tree model for our line classification. The experimental results are explained in the Experiments section. 

\subsubsection*{Naive Bayes(NB)}
Given a dependent feature vector set, $F$ $=$ \{$f_1$,\; $f_2$,\;....,\;$f_n$\} for each line of text from a set of text lines, $T$ $=$ \{$x_1$,\; $x_2$,\;....,\;$x_n$\} and a class label set, $L$ $=$ \{$0$,\;$1$\}, we can calculate the probability of each class $c_i$ from $L$ using the Bayes theorem states in equation \ref{eq:bayes_1}.

\begin{equation}
\label{eq:bayes_1}
  P(c_i|F) = \frac{P(c_i)\;.\;P(F|c_i)}{P(F)}
\end{equation}

As $P$($F$) is the same for the given input text, we can determine the class label of a text line having feature vector set $F$, using the equation \ref{eq:bayes_2}.

\begin{equation}
\label{eq:bayes_2}
   \begin{rcases} Label(F) = arg\;Max_{c_i}\{P(c_i|F)\} \\
  		   \;\;\;\;\;\;\;\;\;\;\;\;\;\, = arg\;Max_{c_i}\{P(c_i)\;.\;P(F|c_i)\}	
    \end{rcases}
\end{equation}

Here, the probability \begin{math} P(F|c_i) \end{math} is calculated using the multinomial Naive Bayes method. 
We use multinomial Naive Bayes method from scikit-learn \cite{scikit-learn} to train models, where the feature vector, $F$ is either $16$ features from layout or $16$ layout features concatenated with the word vector of the text line.

\subsubsection*{Recurrent Neural Networks(RNN)}
\begin{figure}
\includegraphics[height=0.7in, width=2.5in]{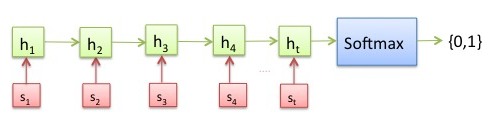}
\caption{Many-to-one RNN approach for line classification\label{fig:many_to_one_RNN_more_detail}}
\end{figure}

Given an input sequence, $S$ $=$ \{$s_1$,\; $s_2$,\;....,\;$s_t$\} of a line of text, we train a character level RNN model to predict it's label, \begin{math}
$l$ \;\in\; $L$ = \{ \textit{regular-text}:$0$, \; \textit{section-header}:$1$\}
\end{math}. We use a many-to-one RNN approach, which reads a sequence of characters until it gets the \textit{end of the sequence} character. It then predicts the class label of the sequence. The RNN model takes the embeddings of characters in the text sequence as input. For character embedding, we represent the sequence into a character level one-hot matrix, which is given as input to the RNN network. It is able to process the sequence recursively by applying a transition function to it's hidden unit, $h_t$. The activation of the hidden unit is computed by the equation \ref{eq:rnn_actionation}.

\begin{equation}
\label{eq:rnn_actionation}
  h_t= \begin{cases}
  	0 \;\;\;\;\;\;\;\;\;\;\;\;\;\;\; \;\;\;\;\; t=0 \\
  	f(h_{t-1}, s_t) \;\;\;\;\;  otherwise
	\end{cases}
\end{equation}

where $h_t$ and $h_{t-1}$ are the hidden units at time $t$ and $t-1$ and $s_t$ is the input sequence from the text line at time $t$. 
The RNN maps the whole sequence of characters until the \textit{end of the sequence} character with a continuous vector, which is input to the $softmax$ layer for label classification. A many-to-one RNN architecture for our line classification is shown in figure \ref{fig:many_to_one_RNN_more_detail}. 

\begin{figure*}
\includegraphics[height=2.0in, width=5.0in]{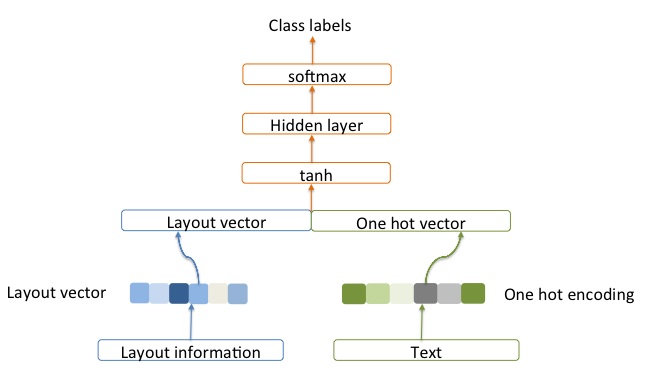}
\caption{RNN architecture for layout and text\label{fig:full_architecture_rnn}}
\end{figure*}

We use Tensorflow \cite{abadi2016tensorflow} to build our RNN models. We build three different networks for our line classification task. In the first and second networks, we use only text and layout as input sequence respectively.  In the third network, we use both 16 layout features and the text as input, where the one-hot matrix of characters sequence is concatenated at the end of the layout features vector. Finally, the whole vector is given as input to the network. Figure \ref{fig:full_architecture_rnn} shows the complete network architecture for layout and text input. The implementation detail is given in the Experiments section.  

\subsubsection{Section Classification}
The section classification module identifies different levels of section headers such as \textit{top-level section}, \textit{subsection} and \textit{sub-subsection} headers. It also detects section boundaries. It has Section Classifiers module and Section Boundary Detector component, which are explained below. 

\subsubsection*{Section Classifiers}
\begin{figure}
\includegraphics[height=0.9in, width=3.0in]{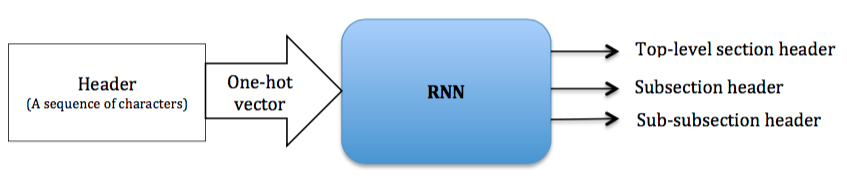}
\caption{Input-output for section classification\label{fig:rnn_for_section_classifier}}
\end{figure}

Like as the Line Classifiers module, the Section Classifiers module considers the section classification task as a prediction modeling problem where we have sequence of inputs $S$ $=$ \{$s_1$,\; $s_2$,\;....,\;$s_t$\} from a classified section header and the task is to predict a category from $L$ = \{ \textit{top-level\; section \;header}:$1$, \; \textit{ subsection \;header}:$2$ \; \textit{sub-subsection \;header}:$3$\} for the sequence. For this sequence prediction task, we use an RNN architecture similar to the architecture used for the line classification. The differences are input sequence and the class labels. The input and output of RNN for this task is shown in figure \ref{fig:rnn_for_section_classifier}. 

\subsubsection*{Section Boundary Detector}
\begin{figure}
\includegraphics[height=1.8in, width=3.2in]{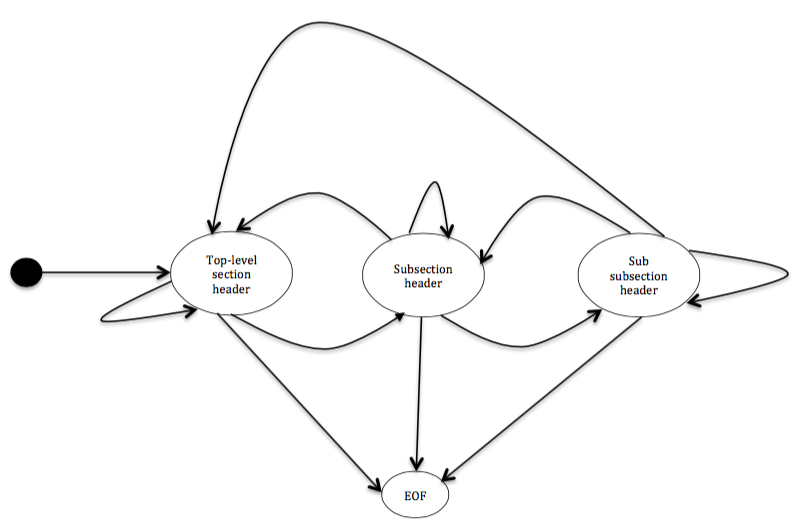}
\caption{Top-level section, subsection and sub-subsection header dependency sequence\label{fig:state_diagram_toc}}
\end{figure}

After identifying different level section headers, we merge all contents (regular text, top-level section header, subsection header and sub-subsection header) with their class labels in a sequential order as they appear in the original document. The Section Boundary Detector splits the whole document into different sections, subsection and sub-subsections based on the given splitting level. By default, it splits the document into top-level sections. It returns output as a dictionary where the keys are text, title and subsections for each section. The subsection has the similar nested structure. The Section Boundary Detector finds the relationship among sections, subsections and sub-subsections using the dependency state diagram presented in figure \ref{fig:state_diagram_toc}. The high level algorithm to generate sections, subsections and sub-subsections using the dependency diagram and class labels is presented in algorithm \ref{alg:sectionboundary}.  

\begin{algorithm}
	\caption{Section boundary detector}\label{alg:sectionboundary}
	\begin{algorithmic}[1]
		\Procedure{split\_doc\_into\_sections}{$doc,split\_level$}
        \State sections =[]
		\If{$split\_level$ is top\_level}
        	\For{$line$ in $doc$}
            \State Generate text\_block based on $class\_label$ = $1$
            \State Add \{title, text\_block\} in sections
            \EndFor
		\ElsIf{$split\_level$ is subsection}
        	\For{$line$ in $doc$}
            	\State Generate text\_block based on $class\_label$ =$1$
            \EndFor
            \For{$block$ in $text\_block$}
               	\State Generate sub\_block based on $class\_label$ =$2$
                \State Add \{title, sub\_block\} in sections
            \EndFor
         \Else
           \For{$line$ in $doc$}
            	\State Generate text\_block based on $class\_label$ =$1$
           \EndFor
           \For{$block$ in $text\_block$}
                \State Generate sub\_block based on $class\_label$ =$2$
           \EndFor
		   \For{$block$ in $sub\_block$}
               	\State Generate sub\_sub\_block based on $class\_label$ =$3$
                \State Add \{title, sub\_sub\_block\} in sections
           \EndFor
		\EndIf
		\State \textbf{return} $sections$
		\EndProcedure
	\end{algorithmic}
\end{algorithm}

\subsubsection{Semantic Annotation}
Given a set of physically divided sections $D$ $=$ \{$d_1$,\; $d_2$,\;....,\;$d_n$\}, the semantic annotation module assigns a human understandable semantic name to each section. We use Latent Dirichlet Allocation (LDA) \cite{blei2003latent} to find a semantic concept from a section. LDA is a generative topic model, which is used to understand the hidden structure of a collection of documents. In LDA, each document has a mixture of various topics with a probability distribution. Again, each topic is a distribution of words. 

Using Gensim\footnote{https://radimrehurek.com/gensim/models/ldamodel.html}, we train an LDA topic model on a set of divided sections. The model is used to predict the topic for any test section. A couple of terms having the highest probability values of the predicted topic are used to annotate the section as a semantic label. 

Using the Section Boundary Detector from Section Classification sub unit, the Semantic Annotation module generates a table of contents (TOC) for any PDF document. It also summarizes each section of a PDF document using the textrank algorithm \cite{mihalcea2004textrank} implemented in NLTK \footnote{http://www.nltk.org/}, where sections are detected by the Section Boundary Detector. 

\section{Experiments and Evaluation of Results}
We evaluated the effectiveness of our approaches using scientific articles from arXiv\footnote{https://arxiv.org} repository. This section describes data, experiments and evaluation of our results.

\subsection{Data Construction}
\subsubsection{Data Collection}
We downloaded all \textit{arXiv} articles from Amazon $S3$ cloud storage using arXiv Bulk Data Access option uploaded by arXiv for the time period of $2010$ to $2016$ December. The files were grouped into .tar files of \begin{math}\sim 500MB\end{math} each. The total size of all files is $743.4GB$. After downloading, we extracted all tar files and got $1121363$ articles in PDF. Using open archives initiative (OAI)\footnote{http://www.openarchives.org/OAI/2.0/openarchivesprotocol.htm} protocol, we harvested metadata for each of the articles from \textit{arXiv} repository. The metadata includes title, publication date, abstract, categories and author names. Some of the \textit{arXiv} articles have bookmarks. We also extracted bookmarks from each article. We kept the hierarchy in the bookmarks. We considered bookmarks as the table of contents(TOC). We combined metadata, the TOC and a downloadable link for each article and stored in a JSON file where \textit{arXiv} file name is the key for each set of information. 

\subsubsection{Data Processing}
We converted each PDF article to an XML dialect called TETML\footnote{Text Extraction Toolkit Markup Language} using PDFLib. The granularity of the conversion was word level. After conversion, the total size of all TETML files was 5.1TB. The elements are organized in a hierarchical order in a TETML file. Each TETML file contains pages. Each page has annotation and content elements. The content element has all of the text blocks in a page as a list of para elements. Each para element has a list of words where each word contains a high level description of each character such as font name, size, weight, x-y coordinates and character width. Our parser reads the structure of the TETML file and parses it. The parser processes a description of each character and generates text lines and layout information from the description for each line by applying different heuristics. The layout information are the starting and ending of x and y positions of a line, font size, font weight, font-family, page number, page width and page height. It returns all lines of text with layout information.
\subsubsection{Training and Test Data}
For our experiments on arXiv articles, we have a component, which processes bookmarks and each TETML file. After getting all lines of text with layout information from the parser, the component traverses the TOC for each file and maps each element of the TOC with text lines from the document. It finds a path for each element of the TOC and defines a class label for each line based on the mapping between the TOC element and text line. The class labels are regular-text:0, top-level section header:1, subsection header:2 and sub-subsection header:3.  Finally, we generated a dataset in a CSV format where each row has text line, layout information, file name of that line and class label of that line. This dataset is used as gold standard data for our experiments. We took 60\%  as training and 40\% as test out of $1121363$ articles which have TOCs. Our developed models identify sections and the TOCs for the rest of the data. 

\subsection{Experiment for Line Classification}
As explained in the approach section, we used SVM, Decision Tree, Naive Bayes and RNN classifiers for our line classification. Table \ref{tab:classifier_configuration} shows the configurations of our classifiers. As a document has very few section headers with respect to regular text, our data is highly imbalanced and some of the layout features depend on the sequence of lines. After generating features, we balanced our dataset. We considered an equal number of samples for all the classes. As the \textit{arXiv} dataset is very large, we only took a part of the dataset to train and test our models. Table \ref{tab:training_test_data_size} shows the training and test dataset size for our experiments.

\begin{table*}
\setlength\abovecaptionskip{3pt}
  \caption{Classifiers configurations}
  \label{tab:classifier_configuration}
  \begin{tabular}{llll}
    \toprule
    SVM & DT & NB & RNN\\
    \midrule 
    \shortstack[l]{kernel=’linear’ \\
     regularization = 'l2' \\
     features = 'layout', 'layout and text' \\
     vectorizer = TF-IDF vectorizer \\
     ngram= unigram, bigram and trigram\\
     minimum doc frequency = 5\% \\
     maximum doc frequency = 95\%}
     &
     \shortstack[l]{criterion = 'gini'\\
     algorithm = 'CART'\\
     features = 'layout', 'layout and text' \\
     vectorizer = TF-IDF vectorizer \\
     ngram= unigram, bigram and trigram\\
     minimum doc frequency = 5\% \\
     maximum doc frequency = 95\%  }
  	 &
     \shortstack[l]{algorithm ='MultinomialNB'\\
     features = 'layout', 'layout and text' \\
     vectorizer = TF-IDF vectorizer \\
     ngram= unigram, bigram and trigram\\
     minimum doc frequency = 5\% \\
     maximum doc frequency = 95\% }
     &
	\shortstack[l]{
    max\_doc\_len = 100\\
    hidden\_size = 20\\
    encoding = 'one-hot'\\
    optimizer = 'adam'\\
    learning\_rate =0.001\\
    function = 'Softmax' \\
    batch\_size = 10 
    }
     \\
  \bottomrule
\end{tabular}
\end{table*}


\begin{table}
\setlength\abovecaptionskip{3pt}
  \caption{Training and Test Data for Line Classification}
  \label{tab:training_test_data_size}
  
  \begin{tabular}{lcc}
	\toprule
	\multirow{3}{*}{} & Training Data & Test Data\\
    \midrule 
 	Regular-Text & 121077 & 80184 \\
 	Section-Header & 121077 & 80184  \\
    Top-level Section Header & 208430 & 166744  \\
    Subsection Header & 208430 & 166744  \\
    Sub-subsection Header & 208430 & 166744  \\
	\bottomrule
	\end{tabular}

\end{table}

\begin{figure}
\includegraphics[height=1.7in, width=3.5in]{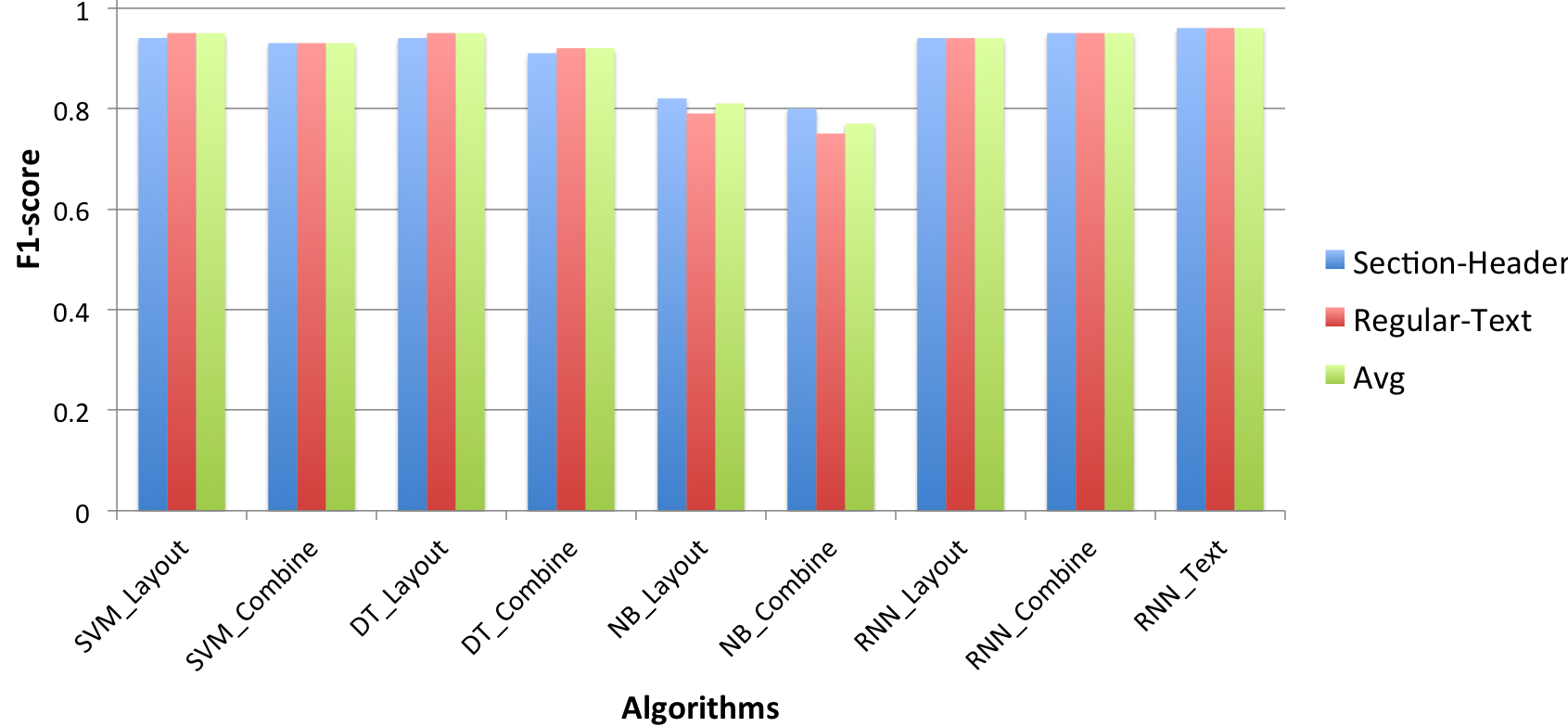}
\caption{Performance Comparison for line classification \label{fig:comparison_f1_score_all_algorithms}}
\end{figure}
\begin{figure*}
\begin{subfigure}{.25\textwidth}
  \centering
  \includegraphics[height=1.0in, width=1.8in]{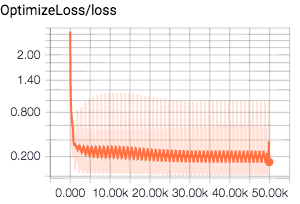}
  \caption{Layout}
  \label{fig:loss_layout}
\end{subfigure}%
\begin{subfigure}{.24\textwidth}
  \centering
  \includegraphics[height=1.0in, width=1.8in]{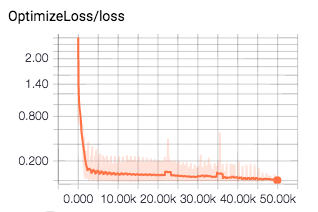}
  \caption{Text}
  \label{fig:loss_text}
\end{subfigure}%
\begin{subfigure}{.25\textwidth}
  \centering
  \includegraphics[height=1.0in, width=1.8in]{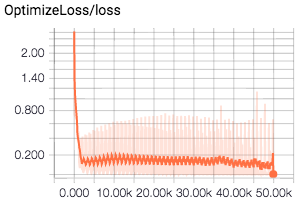}
  \caption{Combine}
  \label{fig:loss_combine}
\end{subfigure}%
\begin{subfigure}{.25\textwidth}
  \centering
  \includegraphics[height=1.0in, width=1.8in]{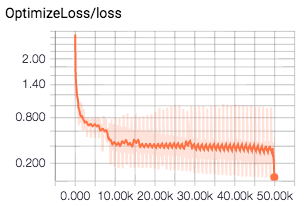}
  \caption{Sections}
  \label{fig:loss_sections}
\end{subfigure}

\caption{Training Loss\label{fig:all_loss}}
\end{figure*}

To evaluate our models, we used precision, recall and f-measure.  Table \ref{tab:p_r_f1_all_models_both_layout_combine_features} shows precision, recall and f1 scores for all of our approaches on the test dataset. We also trained a character level RNN model using only the text. Precision, recall and f1 scores for this model are shown in table \ref{tab:text_features_rnn_line_classification}. Figure \ref{fig:comparison_f1_score_all_algorithms} compares f1 scores for all of the algorithms we used for line classification. We achieved the best performance with character level RNN using only text as input. Figure \ref{fig:loss_layout}, \ref{fig:loss_text} and \ref{fig:loss_combine} show the training losses over the number of steps for RNN with layout, text and combine input respectively where we got minimum loss for text input. 

\begin{table}
\setlength\abovecaptionskip{3pt}
\caption{For both Layout and Combine Features}
\label{tab:p_r_f1_all_models_both_layout_combine_features}
\resizebox{\columnwidth}{!}{
\begin{tabular}{llcccccc}
\toprule
Algorithms           & \multicolumn{4}{c}{Layout Features}            & \multicolumn{3}{c}{Combine Features} \\
\multirow{3}{*}{SVM} &                & Precision & Recall & F1 Score & Precision    & Recall   & F1 Score   \\
\midrule 
                     & Section-Header & 0.97         & 0.92      & 0.94        & 0.93            & 0.92        & 0.93          \\
                     & Regular-Text   & 0.93         & 0.97      & 0.95        & 0.92            & 0.93        & 0.93          \\
\midrule                      
\multirow{2}{*}{DT}  & Section-Header & 0.97         & 0.92      & 0.94        & 0.96            & 0.87        & 0.91          \\
                     & Regular-Text   & 0.92         & 0.97      & 0.95        & 0.88            & 0.97        & 0.92          \\
\midrule                      
\multirow{2}{*}{NB}  & Section-Header & 0.76         & 0.90      & 0.82        & 0.73            & 0.89        & 0.80          \\
                     & Regular-Text   & 0.88         & 0.72      & 0.79        & 0.85            & 0.67        & 0.75          \\
\midrule                      
\multirow{2}{*}{RNN} & Section-Header & 0.94         & 0.94      & 0.94        & 0.95            & 0.95        & 0.95          \\
                     & Regular-Text   & 0.94         & 0.94      & 0.94        & 0.95            & 0.95        & 0.95          \\
\bottomrule
\end{tabular}}
\end{table}

\begin{table}
\setlength\abovecaptionskip{3pt}
  \caption{Text only using RNN for Line Classification}
  \label{tab:text_features_rnn_line_classification}
  
  \begin{tabular}{lccc}
  \toprule
	\multicolumn{1}{c}{} & Precision & Recall & F1 Score \\
    \midrule                      
	Section-Header       & 0.97        & 0.96      & 0.96        \\
	Regular-Text         & 0.95         & 0.97      & 0.96       \\
   
   \bottomrule
   \end{tabular}
\end{table}

\subsection{Experiment for Section Classification}
As we achieved the best result for line classification using the RNN model, we chose RNN for section classification. We also prepared a training and test dataset for this task. Table \ref{tab:training_test_data_size} shows the size of training and test datasets for section classification. Precision, recall and F1 scores for section classification are shown in table \ref{tab:p_r_f1_rnn_section_classification}. From figure \ref{fig:loss_sections}, we can see that the training loss is higher in sections classification than line classification. It is obvious that identifying \textit{top-level}, \textit{subsection} and \textit{sub-subsection} headers are more complex than just identifying \textit{section-header} or \textit{regular-text}.  

\begin{figure*}
\includegraphics[width=\textwidth,height=\textheight,keepaspectratio]{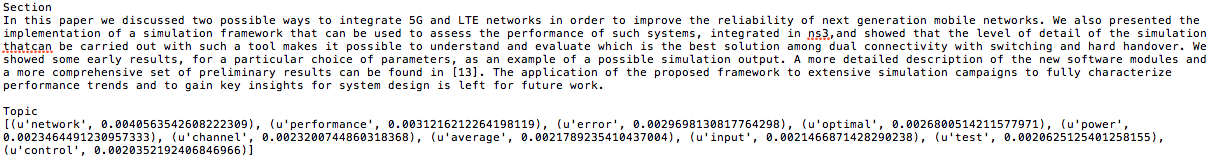}
\caption{Semantic Annotation using top terms from LDA topic \label{fig:lda_topic_annotation}}
\end{figure*}

\begin{figure*}
  \includegraphics[width=\textwidth,height=0.8cm]{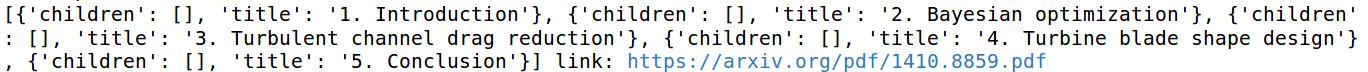}
  \caption{Only top-level section headers}
  \label{fig:Only_top-level_section_headers}
\end{figure*}

\begin{figure*}
  \includegraphics[width=\textwidth,height=1.0cm]{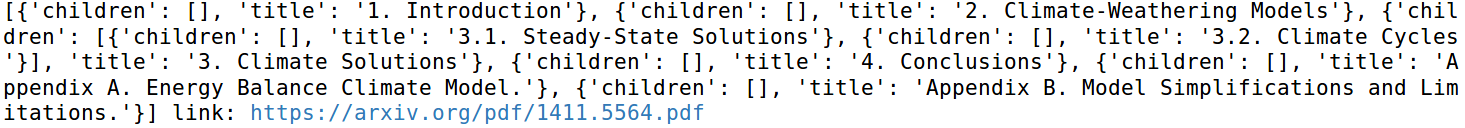}
  \caption{Top-level and subsection headers}
  \label{fig:Top-level_and_subsection_headers}
\end{figure*}

\begin{table}
\setlength\abovecaptionskip{3pt}
  \caption{For Section Classification using RNN}
  \label{tab:p_r_f1_rnn_section_classification}
  \begin{tabular}{lccc}
  \toprule
	\multicolumn{1}{c}{} & Precision & Recall & F1 Score \\
    \midrule                      
	Top-level Section Header       & 0.83        & 0.88      & 0.85        \\
	Subsection Header         & 0.81         & 0.81      & 0.81       \\
    Sub-subsection Header         & 0.78         & 0.73      & 0.75       \\
    \midrule                      
    Avg         & 0.81         & 0.81      & 0.81       \\
   \bottomrule
   \end{tabular}
\end{table}

\begin{figure*}
\includegraphics[height=3.5in, width=6.5in]{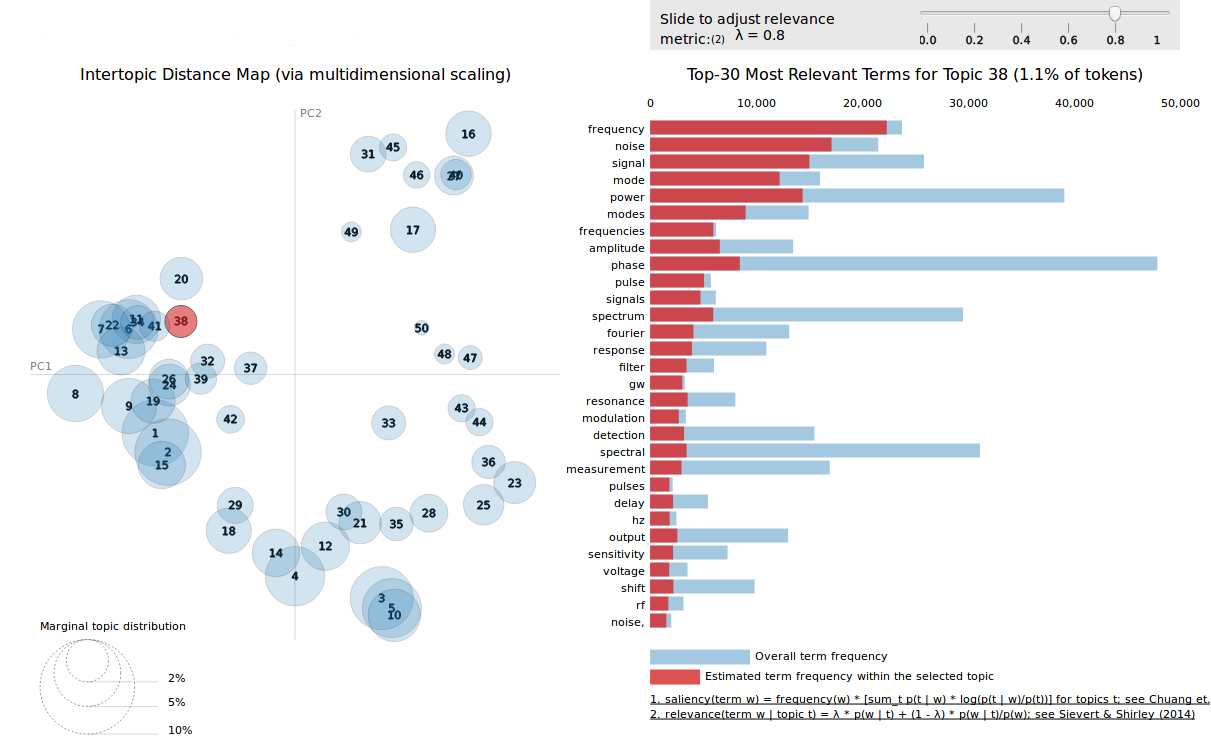}
\caption{Inter topic distance map and top terms for a topic \label{fig:lda_inter_topics_distance_terms}}
\end{figure*}

\begin{figure}
\includegraphics[height=1.5in, width=3.0in]{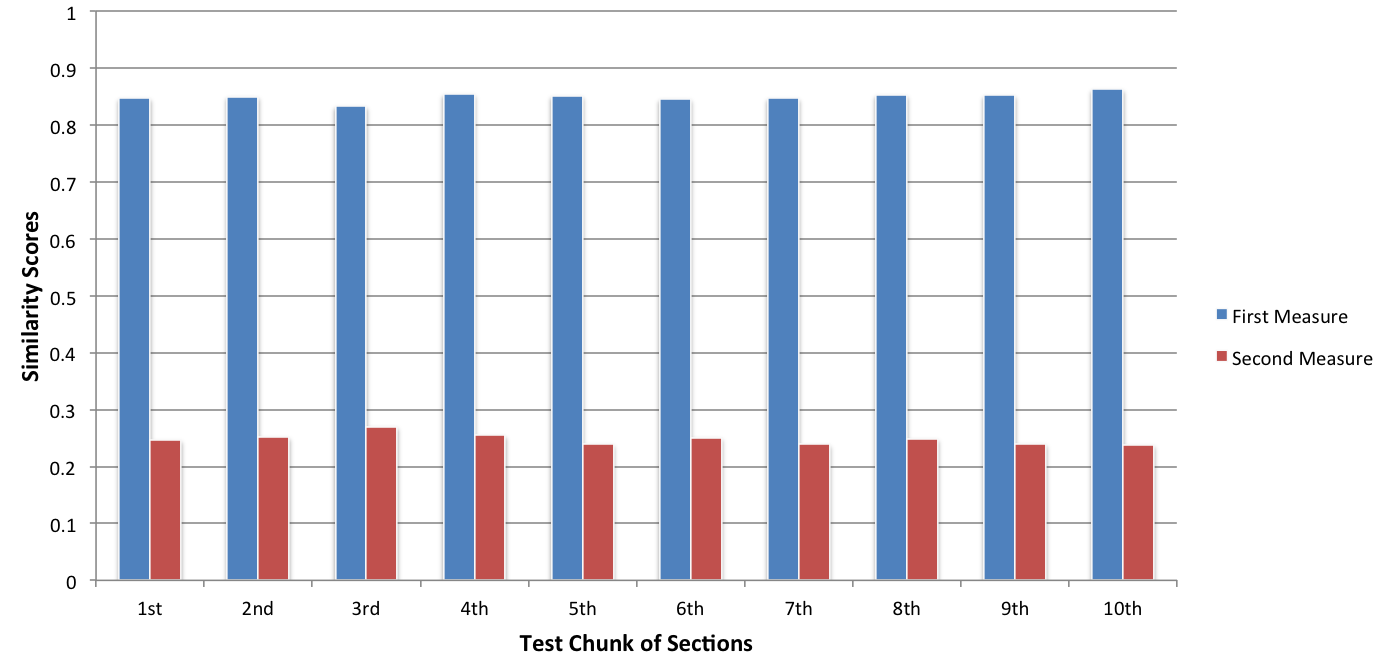}
\caption{Similarity measures for LDA\label{fig:lda_similarity_measure_two}}
\vspace{-1.5em}
\end{figure}

\subsection{Experiment for Semantic Annotation}
We trained an LDA model on $128505$ divided sections through 50 passes for a different number of topics and evaluated the model on $11633$ divided sections. While building the dictionary for the model, we ignored words that appear in less than 20 sections or more than 10\% of sections. Our final dictionary size after filtering was $100000$. Figure \ref{fig:lda_inter_topics_distance_terms} shows inter topic distance map for 10 topics where some of the topics overlap. This figure also shows the 30 most relevant terms for topic 4 where the relevance score is 80\%. To annotate a section, we used the model to get the best topic for that section and chose a couple of terms with the highest probability. An example is shown in figure \ref{fig:lda_topic_annotation}. To evaluate the LDA model for sections, we considered perplexity and cosine similarity measures. The perplexity for test chunk is -$9.684$ for 10 topics. The perplexity is lower in magnitude which means that the LDA model fits better for the test sections and probability distribution is good at predicting the sections. We split the test set into 10 different chunks of test sections where each chunk has $1000$ sections without repetition. We also split each section from each test chunk into two parts and checked two measures. The first measure is similarity between topics of the first half and topics of the second half for the same section. The second measure is similarity between halves of two different sections. We calculated average cosine similarity between parts for each test chunk of sections. Due to the coherence between topics, the first measure should be higher and the second measure should be lower. Figure \ref{fig:lda_similarity_measure_two} shows these two measures for 10 different chunk of test sections. We also generated TOCs from any scholarly article. Figure \ref{fig:Only_top-level_section_headers} and \ref{fig:Top-level_and_subsection_headers} show the TOCs from two different articles where each TOC represents the hierarchies of different section headers.

\subsection{Comparison of Results and Discussion}
We compared the performance of our framework in the previous sections with respect to different performance matrices. We also compared the performance of our framework against the top performing systems for scholarly articles in PDF form. The first comparison system is PDFX presented by Alexandru Constantin et al. in \cite{constantin2013pdfx}. Our task is formalized in a different way and partially similar to their task. Their system identifies author, title, email, section headers etc. from scholarly articles. They reported an f1 score of 77.45\% for top-level section headers identifying for a various articles. The dataset is not publicly available. We achieved an 85\% f1 score for top-level section headers identifying along with a 96\% f1 score for just section header identifying from \textit{arXiv} repository which has various types of academic articles from thousands of different categories and subcategories. The second comparison system is a hybrid approach to discover semantic hierarchical sections from scholarly documents by Suppawong Tuarob et al. \cite{tuarob2015hybrid}. Their task is limited to a few fixed section heading name identifications whereas our framework can identify any heading name. Their dataset is not directly applicable to our system, but it is on scholarly articles. They got a 92.38\% f1 score for section boundary detection where sections are of any level(from fixed names such as abstract, introduction and conclusions) and we got a 96\% f1 score for any heading name identification. We also tried our framework on business documents such as a Request for Proposal (RFP) dataset collected from a startup company that works on business documents analysis. RFPs are usually large, complex and very unstructured documents. Due to the terms and conditions given by the company, we are not able to present results and that dataset in this research paper. 

As we use PDFLib for PDF extraction, we depend on their system performance. Due to the different encoding of PDF documents, sometimes PDFLib identifies text block incorrectly and classifies a same block into two different blocks. This generates an error in our data when we map bookmarks in the original PDF for training and test data generation. To reduce this error, we used SequenceMatcher to calculate string similarity score. If the score is more than a threshold, we map the bookmark entry with a line of text from the original PDF. Due to the use of similarity score and threshold heuristic, we may still miss a few section headers. But the ratio is very low. We expect to overcome this error completely in our future work. 

A complete dataset \cite{arxiv_metadata} is available with metadata including a table of contents, section labels, section summarizations, publication history, author names and downloadable \textit{arXiv} link for each article from 1986 to 2016. 

\section{Conclusions and Future work}
We presented a novel framework to understand academic scholarly articles by automatically identifying and classifying sections and labeling them with human understandable semantic names. We experimented with different machine learning approaches and found that RNN works better. We also contributed to the community by releasing a large dataset from scholarly articles. For future work, we plan to develop an ontology to map semantic sections with standard names in different domains. We are also interested in developing a deep learning summarization technique for individual section summarization. Another interesting direction would be to develop an algorithm which can understand any new structure of a large document.

\section{Acknowledgments}
The work presented in this paper was supported by an grant number 1549697 from the National Science Foundation.
